\documentclass{article}
\usepackage{amsmath,epsfig}
\usepackage[preprint]{spconfa4}
\usepackage{xcolor}
\usepackage{cite}
\usepackage[hyphens]{url}  % DO NOT CHANGE THIS
\usepackage{graphicx} % DO NOT CHANGE THIS
\usepackage{booktabs}
\usepackage{amssymb}
\usepackage{subcaption}
\usepackage{booktabs}
\usepackage{multirow}

\usepackage[colorlinks=true,
citecolor=green,
linkcolor=black,
anchorcolor=black,
urlcolor=blue]{hyperref}

\newcommand{\eat}[1]{}

\newcommand{\eg}{\textit{e.g., }}

\let\OLDthebibliography\thebibliography
\renewcommand\thebibliography[1]{
  \OLDthebibliography{#1}
  \setlength{\parskip}{0pt}
  \setlength{\itemsep}{0pt plus 0.3ex}
}

\begin{document}\sloppy

% Example definitions.
% --------------------
\def\x{{\mathbf x}}
\def\L{{\cal L}}

% Title.
% ------
\title{X-HRNet: Towards Lightweight Human Pose Estimation with Spatially Unidimensional Self-Attention}
%
% Address.
% ---------------
\name{Yixuan Zhou, Xuanhan Wang, Xing Xu$^*$, Lei Zhao and Jingkuan Song
\thanks{$^*$Corresponding author: Xing Xu. This work was supported in part by National Natural Science Foundation of China (No.61976049 and No.62072080) and Sichuan Science and Technology Program, China (No. 2019ZDZX0008, 2019YFG0533).}
}
\address{Center for Future Media \& School of Computer Science and Engineering \\
University of Electronic Science and Technology of China, China}

\maketitle

\begin{abstract}
High-resolution representation is necessary for human pose estimation to achieve high performance, and the ensuing problem is high computational complexity.
In particular, predominant pose estimation methods estimate human joints by 2D single-peak heatmaps. Each 2D heatmap can be horizontally and vertically projected to and reconstructed by a pair of 1D heat vectors.
Inspired by this observation, we introduce a lightweight and powerful alternative, \textbf{S}patially \textbf{U}nidimensional \textbf{S}elf-\textbf{A}ttention (\textbf{SUSA}), to the pointwise ($1\times1$) convolution that is the main computational bottleneck in the depthwise separable $3\times3$ convolution.
Our SUSA reduces the computational complexity of the pointwise ($1\times1$) convolution by $96\%$ without sacrificing accuracy.
Furthermore, we use the SUSA as the main module to build our lightweight pose estimation backbone X-HRNet, where $X$ represents the estimated cross-shape attention vectors. 
Extensive experiments on the COCO benchmark demonstrate the superiority of our X-HRNet, and comprehensive ablation studies show the effectiveness of the SUSA modules.
The code is publicly available at \url{https://github.com/cool-xuan/x-hrnet}.
\end{abstract}
\section{Introduction}
\label{sec:intro}
2D human pose estimation aims to localize all human joints from the whole image space. Recent works \cite{badrinarayanan2017segnet, chen2014semantic, sun2019deep, wang2020deep, cheng2020higherhrnet, li2020simple} demonstrate that the high-resolution representation is necessary for this pixel-level regression task to achieve high performance.
But massive computational resources are required in these high-resolution models, and it is hard to deploy them on the increasingly popular mobile devices with limited resources. Therefore, building a lightweight and efficient pose estimator has become an urgent challenge.

\begin{figure}
  \setlength{\belowcaptionskip}{-0.3cm}
  \centering
  \begin{subfigure}[b]{0.28\linewidth}
      \centering
      \includegraphics[width=\textwidth]{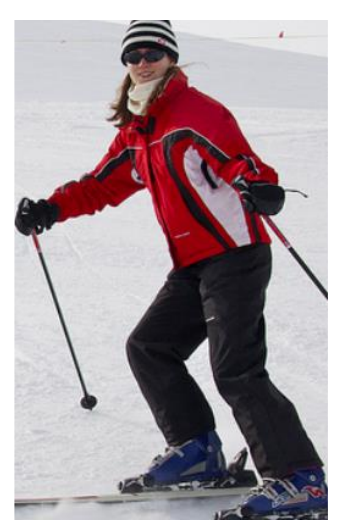}
      \caption{}
      \label{fig:introduction-img}
  \end{subfigure}
  \hfill
  \begin{subfigure}[b]{0.28\linewidth}
      \centering
      \includegraphics[width=\textwidth]{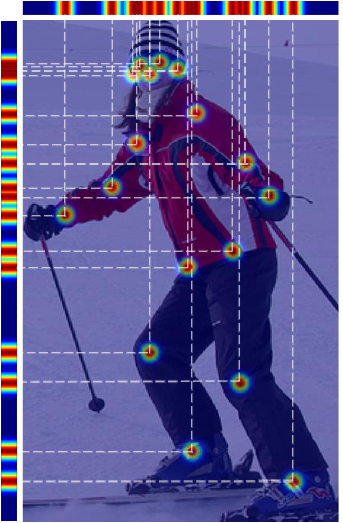}
      \caption{}
      \label{fig:introduction-pose}
  \end{subfigure}
  \hfill
  \begin{subfigure}[b]{0.32\linewidth}
      \centering
      \includegraphics[width=\textwidth]{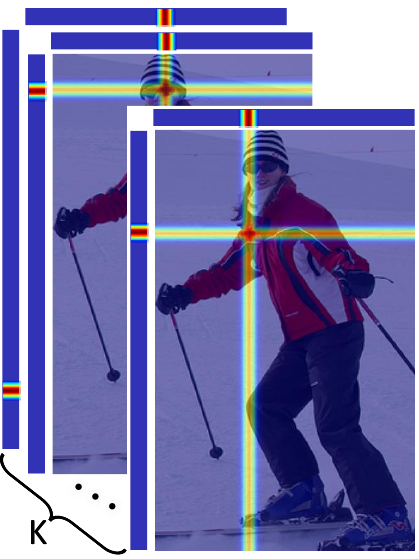}
      \caption{}
      \label{fig:introduction-Reconstruction}
  \end{subfigure}
     \caption{(a) Input image. (b) 2D confidence heatmap can be horizontally and vertically projected to two 1D vectors. (c) $K$ pairs of 1D heat vectors reconstruct $K$ 2D heatmaps.}
  \label{fig:introduction-idea}
\end{figure}

\begin{figure*}
    \setlength{\belowcaptionskip}{-0.3cm}
    \centering
    \begin{subfigure}[b]{0.211\linewidth}
        \centering
        \includegraphics[width=\textwidth]{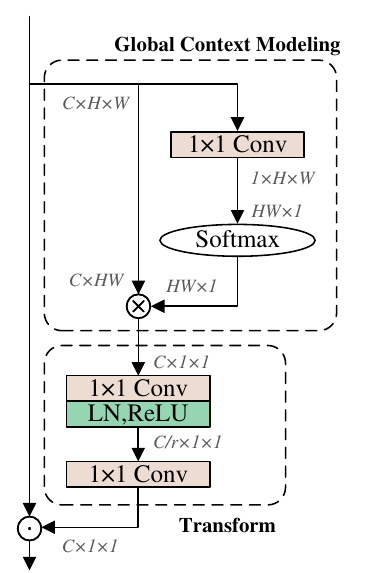}
        \caption{GC Block}
        \label{fig:method-gbblock}
    \end{subfigure}
    \hfill
    \begin{subfigure}[b]{0.255\linewidth}
        \centering
        \includegraphics[width=\textwidth]{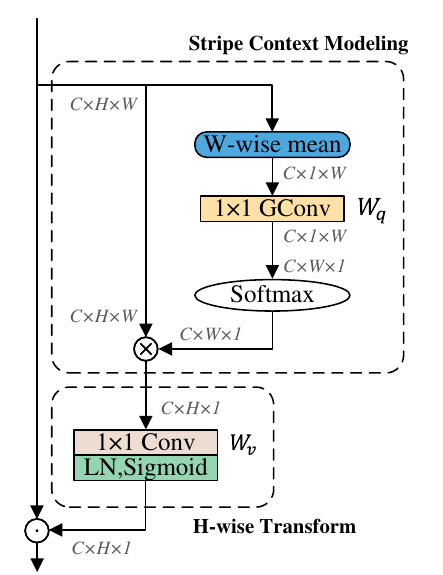}
        \caption{H-wise SUSA}
        \label{fig:method-hwisesusa}
    \end{subfigure}
    \hfill
    \begin{subfigure}[b]{0.255\linewidth}
        \centering
        \includegraphics[width=\textwidth]{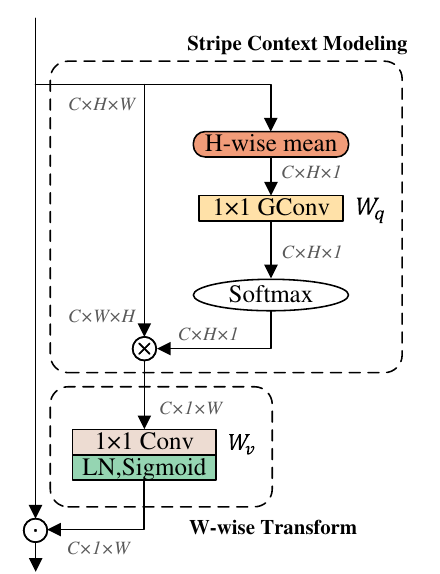}
        \caption{W-wise SUSA}
        \label{fig:method-wwisesusa}
    \end{subfigure}
    \hfill
    \begin{subfigure}[b]{0.22\linewidth}
        \centering
        \includegraphics[width=0.86\textwidth]{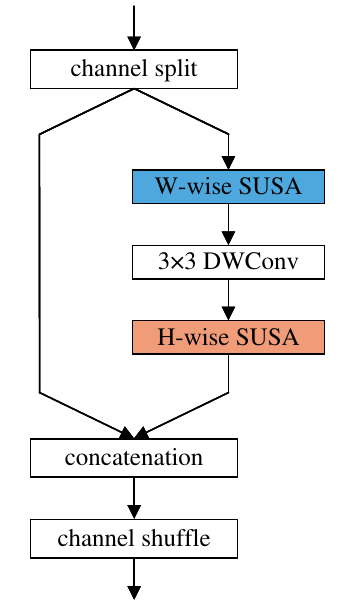}
        \caption{X-shuffle block}
        \label{fig:method-x-shuffle-block}
    \end{subfigure}
    \caption{\textbf{Architecture of the GC block \cite{cao2019gcnet}, our SUSA and X-shuffle block.} For intuitive understanding, the features are abstracted as feature dimension, \eg $C \times H \times W$ denotes a feature map with channel number $C$, height $H$ and Width $W$. $\otimes$ denotes matrix inner-product, $\odot$ denotes element-wise multiplication, and $\oplus$ denotes element-wise addition.}
    \label{fig:method}
  \end{figure*}

There are various lightweight schemes \cite{zhang2018shufflenet, ma2018shufflenet, howard2018inverted} for the classification task that can be borrowed from, where depthwise separable convolution \cite{chollet2017xception, howard2017mobilenets} (a depthwise $3\times3$ convolution followed by a pointwise convolution) is widely used to replace the standard $3\times3$ convolution.
This substitution reduces the computational complexity by $89\%$ without a performance drop. However, the pointwise ($1\times1$) convolution is still the main computational bottleneck \cite{yu2021lite}. The straightforward optimization scheme reduces the channel number. For example, ShuffleNetV2 \cite{ma2018shufflenet} alleviates the FLOPs of pointwise ($1\times1$) convolution by only feeding half of the features to convolutions. Nevertheless, these lightweight methods tailored for classification focus on channel dimension, neglecting the computational overhead caused by spatial dimensions that are considerable for pixel-wise regression tasks.

Unlike conventional classification networks where the resolution of features shrinks with the increase of the depth, the networks for pixel-wise regression tasks maintain high-resolution features \cite{yu2018bisenet, wang2020deep, cheng2020higherhrnet} or adopt the encoder-decoder structure \cite{badrinarayanan2017segnet, lin2017feature, xiao2018simple}. For example, the HRNet \cite{sun2019deep} has four parallel branches with different resolutions, whose architecture is shown as follows:
{\small
\setlength\abovedisplayskip{5pt}
\setlength\belowdisplayskip{3pt}
{
\begin{equation}
    \begin{aligned}
    \mathcal{N}_{11} &\longrightarrow &\mathcal{N}_{21} &\longrightarrow &\mathcal{N}_{31} &\longrightarrow &\mathcal{N}_{41} \\
                      & \searrow       &\mathcal{N}_{22} &\longrightarrow &\mathcal{N}_{32} &\longrightarrow &\mathcal{N}_{42} \\
                      &                &                 & \searrow       &\mathcal{N}_{33} &\longrightarrow &\mathcal{N}_{43} \\
                      &                &                 &                &                 &\searrow        &\mathcal{N}_{44}
    \end{aligned}
\end{equation}
}}where $\mathcal{N}_{sb}$ denotes the subnetwork of the $s$ stage in the $b$ branch. The first branch maintains the highest resolution, whose size is a quarter of the one for input images. In HRNet, the width $H$ and the height $W$ of the feature map $x\in \mathbb{R}^{C\times H \times W}$ are on the same order of magnitude as the number of channel $C$. Therefore, we shift the focus from $C$ to spatial dimensions: $H$ and $W$ for reducing the computational complexity in high-resolution pose estimation.
Similar to our scheme, \cite{yu2021lite} presents the Lite-HRNet to reduce the computational complexity in spatial dimension as well.
In the Lite-HRNet, all features over multiple resolutions are downsampled to the minimum resolution and concatenated together. The fused features are fed into the conditional channel weighting (CCW) to estimate the weights conditioned on the input features.
However, high-resolution details are missing severely because of linear interpolation due to simply downsampling features.

To reduce the computational complexity without harming high-resolution details, we explore the essential characteristic of the pose estimation task: maximizing the confidence of all human joint regions in the whole 2D space, and the estimated 2D confidence map can be horizontally and vertically projected to two 1D confidence vectors as illustrated in Figure \ref{fig:introduction-pose}.
Particularly in the predominant top-down pipeline (i.e., detect and estimate) \cite{xiao2018simple, chen2018cascaded, sun2019deep,wang2020deep}, $K$ independent 2D heatmaps are separately estimated to represent the confidence for $K$ human joints. And the 2D heatmap for each joint can be reconstructed by a pair of 1D heat vectors, as shown in Figure \ref{fig:introduction-Reconstruction}.

Inspired by this, we introduce a lightweight unit called \textbf{S}patially \textbf{U}nidimensional \textbf{S}elf-\textbf{A}ttention (\textbf{SUSA}) as an alternative to the computationally costly pointwise ($1\times1$) convolution in high-resolution pose estimation.
It firstly groups the input features along one spatial dimension ($H$ or $W$) by \textbf{S}tripe \textbf{C}ontext \textbf{M}odeling (\textbf{SCM}). 
And then, the grouped features are fed into \textbf{S}patially \textbf{U}nidimensional \textbf{T}ransform (\textbf{SUT}) to learn vertical and horizontal attention vectors. 
It is noteworthy that the SUT only needs to apply convolution on spatially unidimensional features whose computational complexity is linear to the spatial dimension, while the pointwise ($1\times1$) convolution is on quadratic complexity.
We substitute both pointwise ($1\times1$) convolutions in the shuffle block \cite{ma2018shufflenet} and employ the resulting block to build our lightweight high-resolution pose estimation network X-HRNet, where $X$ represents the estimated cross-shape attention vectors.
Without harming the high-resolution details, SUSA is equipped with comparable representation capacity but only requires $2.6\%$ of the FLOPs of the pointwise ($1\times1$) convolution.

Our main contributions can be summarized as follows:
\vspace{-0.8\topsep}
\noindent{
\begin{itemize}
    \setlength{\itemsep}{0pt}
    \setlength{\parsep}{0pt}
    \setlength{\parskip}{3pt}
    \item We introduce a lightweight and efficient module, SUSA, for high-resolution pose estimation. Our SUSA performs calculations over the spatially unidimensional feature instead of the 2D feature, significantly reducing computational complexity.
    \item With the proposed SUSA, we introduce an efficient network, X-HRNet. And X-HRNet achieves the state-of-the-art in terms of complexity and accuracy trade-off on the COCO benchmark.
\end{itemize}}

\section{METHODOLOGY}
In this section, the SUSA module is first introduced and comprehensively compared with the GC block \cite{cao2019gcnet}. Then we use the SUSA module as the main module to build our X-HRNet. 
\subsection{Spatially Unidimensional Self-Attention}
The SUSA module follows the design pattern of the global context block (GC block) \cite{cao2019gcnet} whose detailed structure as illustrated in Figure \ref{fig:method-gbblock}, and it is formed of \textbf{S}tripe \textbf{C}ontext \textbf{M}odeling (\textbf{SCM}) and \textbf{S}patially \textbf{U}nidimensional \textbf{T}ransform (\textbf{SUT}).
In the following subsections, we introduce how SCM groups feature along one spatial dimension and how SUT estimates the attention vector for the orthogonal dimension in the wake of an overview of SUSA. Additionally, we expose the relationship between the GC block and our method.

\vspace{8pt}\noindent\textbf{Overview of SUSA.}
With the input feature map $x\in \mathbb{R}^{C\times H \times W}$, there are two spatial dimensions: $H$ and $W$. Accordingly, we propose two corresponding SUSA: H-wise and W-wise SUSA. As illustrated in Figure \ref{fig:method-hwisesusa} and Figure \ref{fig:method-wwisesusa}, they are exactly identical except for processing different spatial dimensions.
SUSA can be divided into three procedures:
1) \textbf{Stripe context modeling (SCM).} The SCM only groups features along one spatial dimension ($H$ or $W$) with the grouping matrix $x_q$ and outputs stripe context features, unlike global context modeling in GC block that groups the features of all positions together.
2) \textbf{Spatially unidimensional transform (SUT).} The SUT transforms the grouped features via a pointwise ($1\times1$) convolution which learns the attention vector over the remaining spatial dimension.
3) \textbf{Feature Aggregation.} Element-wise multiplication is employed to aggregate the learned attention vector with the input feature map.
Our SUSA is formulated as 
{
\setlength\abovedisplayskip{0pt}
\setlength\belowdisplayskip{3pt}
\begin{equation}
    \begin{aligned}
        & SCM(x) \\
        SUSA(x) = x * SUT(&\overbrace{x_v \times x_q}),
    \end{aligned}
\end{equation}}where $*$ is element-wise multiplication, $\times$ refers to matrix inner-product, and $x_v$ is reshaped from the input feature $x$. We instantiate the H-wise SUSA for introducing details of SCM and SUT, and the W-wise one works on the $W$ dimension in the same way.

\vspace{4pt}\noindent\textbf{Stripe Context Modeling (SCM).}
To achieve the trade-off of presentation capacity and efficiency, we adopt a $1\times1$ group convolution $W_q$ (group=$C$) on $x_w \in \mathbb{R}^{C \times 1 \times W}$ to compute a grouping matrix $x_q$, and the $x_w$ is calculated from $x$ via weighted averaging along $H$ dimension.
The $x_q$ is subsequently activated by Softmax normalization to increase the dynamic range of attention.
The calculation formula of $x_q$ is shown as 
{\setlength\abovedisplayskip{3pt}
\setlength\belowdisplayskip{0pt}
    \begin{equation}
        x_q = \textrm{Softmax}(W_q(\varPhi (x))),
    \end{equation}
}where $\varPhi$ is H-wise weighted averaging. With the grouping matrix $x_q$ and $x_v$ reshaped from $x$, SCM computes the stripe context features $f_h$ by aggregating $x_v$ and $x_q$ together through matrix inner-product.

\vspace{4pt}\noindent\textbf{Spatially Unidimensional Transform (SUT).}
Both CCW \cite{yu2021lite} and GC block use two cascaded $1\times1$ convolutions with bottleneck structure to learn the conditional weights.
This trick reduces the FLOPs but introduces extra convolutions, actually slowing down the inference speed.
For simplification, our H-wise SUT encodes $f_h$ by a single $1\times1$ convolution and outputs the final horizontal attention vector $a_h$.
In particular, $a_h$ is normalized by a LayerNorm (LN) on $C$ dimension like GC block and activated by Sigmoid function. The estimated $a_h$ is broadcast multiplied back to $x$ as the horizontal attention. Correspondingly, the W-wise SUT learns the vertical attention and combines it to $x$ by element-wise multiplication.
The formula of SUT is written as
{
\setlength\abovedisplayskip{2pt}
\setlength\belowdisplayskip{2pt}
\begin{equation}
    a_h = \textrm{Sigmoid}(\textrm{LN}(W_v(f_h))).
\end{equation}}

\begin{figure}
    \setlength{\belowcaptionskip}{-0.3cm}
    \centering
    \begin{subfigure}[b]{0.49\linewidth}
        \centering
        \includegraphics[width=0.55\textwidth]{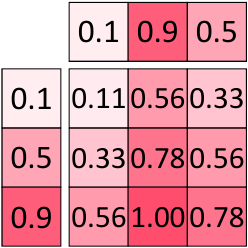}
        \caption{Addition}
        \label{fig:method-Add}
    \end{subfigure}
    \hfill
    \begin{subfigure}[b]{0.49\linewidth}
        \centering
        \includegraphics[width=0.55\textwidth]{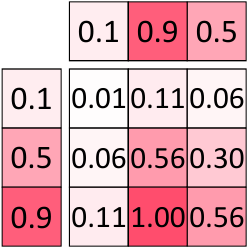}
        \caption{Multiplication}
        \label{fig:method-Mul}
    \end{subfigure}
    \caption{Multiplication fusion produces a sharper peak value and smaller focus region than addition fusion. The output values are normalized after fusion.}
    \label{fig:method-add-mul-fusion}
\end{figure}

\vspace{3pt}\noindent\textbf{Relationship to global context block.} \label{relation-to-gc-block}
Our SUSA module borrows the design scheme from the GC block \cite{cao2019gcnet}. The GC block is an efficient variant of Non-Local Network \cite{wang2018non}, and it aims to capture long-range dependencies in the whole 2D space. We exploit the capacity of capturing long-range dependency into grouping features along one spatial dimension and estimating the stripe context features instead of the global context features. Notably, we aggregate the stripe context features to the input features as horizontal or vertical attention vector by multiplication, while GC Block aggregates global context by addition.
A toy example is shown in Figure \ref{fig:method-add-mul-fusion}, multiplication fusion produces a shaper peak value and smaller activation region than addition fusion. The GC block is designed to model the long-range dependency, and addition fusion learns a large receptive field. However, our SUSA module targets pixel-level peak value maximization, for which multiplication fusion is more suitable.

\subsection{X-HRNet} \label{method-X-HRNet}
To verify the effectiveness of our SUSA modules, we employ them to replace two costly pointwise ($1\times1$) convolutions in the shuffle block \cite{ma2018shufflenet} and name it \textbf{X-shuffle block}, where $X$ represents the estimated cross-shape attention vectors.
In the X-shuffle block shown in Figure \ref{fig:method-x-shuffle-block}, both pointwise ($1\times1$) convolutions are replaced by our H-wise and W-wise SUSA in order. And the order of substitution affects the performance surprisingly.

Based on the X-shuffle blocks, we stack them following the structure of HRNet to maintain the high-resolution representation and build our lightweight backbone \textbf{X-HRNet}.
Particularly, there are three stages in our X-HRNet, and each stage has 2, 3, 4 branches, respectively. The channel dimensions of each resolution branch are $C$, 2$C$, 4$C$, 8$C$, where $C$ is set as 40, following the setting in \cite{yu2021lite}. Besides, all standard $1\times1$ convolutions in fusion modules of HRNet are replaced by depthwise separable $3\times3$ convolutions like \cite{yu2021lite}. We also instantiate X-HRNet-18 and X-HRNet-30 for a fair comparison with \cite{yu2021lite}.
The exhaustive architectures of X-HRNet are illustrated in Table \ref{tab:method-architectures}.
\begin{table}[]
    \setlength{\abovecaptionskip}{0.1cm}
    \setlength{\belowcaptionskip}{-0.3cm}
    \centering
    \caption{\textbf{Architectures of X-HRNet backbones.} 
    $N$ in X-HRNet-$N$ means the number of blocks. $\#$channels indicates the current stage containing multi-branches with a corresponding number of channels.}
    \resizebox{\linewidth}{!}{%
    \begin{tabular}{@{}c|l|l|c|c|ccc@{}}
    \toprule
    \multirow{2}{*}{layer}       & \multirow{2}{*}{block} & \multirow{2}{*}{\#channels} & \multirow{2}{*}{\#blocks} & \multicolumn{2}{c}{\#modules}                                                    \\ \cmidrule(l){5-6}
                                 &                        &                             &                           & \multicolumn{1}{c|}{X-HRNet-18}         & \multicolumn{1}{c}{X-HRNet-30}         \\ \specialrule{.05em}{.25ex}{.25ex}
    image                        &                        & 3                           &                           & \multicolumn{1}{c|}{}                   & \multicolumn{1}{c}{}                   \\ \specialrule{.05em}{.25ex}{.25ex}
    \multirow{2}{*}{stem}        & conv2d                 & 32                          & 1                         & \multicolumn{1}{c|}{\multirow{2}{*}{1}} & \multicolumn{1}{c}{\multirow{2}{*}{1}} \\
                                 & shuffle                & 32                          & 1                         & \multicolumn{1}{c|}{}                   & \multicolumn{1}{c}{}                   \\ \specialrule{.05em}{.25ex}{.25ex}
    \multirow{2}{*}{stage1}      & X-shuffle              & 40,80                       & 2                         & \multicolumn{1}{c|}{\multirow{2}{*}{2}} & \multicolumn{1}{c}{\multirow{2}{*}{3}} \\
                                 & fusion                 & 40,80                       & 1                         & \multicolumn{1}{c|}{}                   & \multicolumn{1}{c}{}                   \\ \specialrule{.05em}{.25ex}{.25ex}
    \multirow{2}{*}{stage2}      & X-shuffle              & 40,80,160                   & 2                         & \multicolumn{1}{c|}{\multirow{2}{*}{4}} & \multicolumn{1}{c}{\multirow{2}{*}{8}} \\
                                 & fusion                 & 40,80,160                   & 1                         & \multicolumn{1}{c|}{}                   & \multicolumn{1}{c}{}                   \\ \specialrule{.05em}{.25ex}{.25ex}
    \multirow{2}{*}{stage3}      & X-shuffle              & 40,80,160,320               & 2                         & \multicolumn{1}{c|}{\multirow{2}{*}{2}} & \multicolumn{1}{c}{\multirow{2}{*}{3}} \\
                                 & fusion                 & 40,80,160,320               & 1                         & \multicolumn{1}{c|}{}                   & \multicolumn{1}{c}{}                   \\ \specialrule{.05em}{.25ex}{.25ex}
    \#Params                     &                        &                             &                           & \multicolumn{1}{c|}{1.3M}               & \multicolumn{1}{c}{2.1M}               \\ \specialrule{.05em}{.25ex}{.25ex}
    \multicolumn{1}{l|}{FLOPs}   &                        &                             & \multicolumn{1}{l|}{}     & \multicolumn{1}{c|}{194.5M}             & \multicolumn{1}{c}{300.4M}             \\ \bottomrule
    \end{tabular}%
    }
    \label{tab:method-architectures}
\end{table}

\section{EXPERIMENTS}
We evaluate our approach on the COCO dataset \cite{lin2014microsoft} and report the comparisons with start-of-the-arts. We also perform comprehensive ablation studies to investigate the effects of each proposed component in our approach.

\subsection{Settings}
\noindent\textbf{Datasets.}
The MS-COCO dataset \cite{lin2014microsoft} includes over 60k images and 250k person instances annotated with 17 human joint categories, \eg left-/right-ear, left-/right-elbow. We train our X-HRNet on the COCO tran2017 dataset, including $57K$ images and $150K$ person instances. The val2017 set and test-dev2017 set are utilized to evaluate our approaches, containing $5K$ and $20K$ images, respectively.

\vspace{4pt}\noindent\textbf{Training.}
We extend the human detection boxes in height or width to a fixed aspect ratio: height : width = 4 : 3, and crop the resized box from the image, which is further rescaled to a fixed size, $256 \times 192$ or $384 \times 288$. Following the data augmentation settings in \cite{yu2021lite}, we randomly augment each person sample with a random scale factor ($[0.75, 1.25]$), random rotation($[-30^{\circ}, 30^{\circ}]$), horizontal flipping and additional half body augmentation.
The network is trained from scratch on 8 NVIDIA V100 with a mini-batch size per GPU. For optimization, we adopt Adam optimizer with an initial learning rate 2$e^{-3}$. The training process is terminated within 210 epochs, and the learning rate is dropped to 2$e^{-4}$ and 2$e^{-5}$ at the 170-th and the 200-th epoch, respectively.

\vspace{4pt}\noindent\textbf{Testing.}
We adopt the top-down paradigm (detect and estimate) with the detection results provided by previous work \cite{xiao2018simple} for a fair comparison. Following the common practice \cite{li2020simple, sun2019deep, wang2020deep}, we compute the final heatmap by averaging the heatmaps of the original and flipped images. The coordinates of keypoints are decoded from the heatmaps by adjusting the highest response location.

\vspace{4pt}\noindent\textbf{Evaluation metric.}
The OKS-based mAP metric is adopted to evaluate our approach. The OKS (Object Keypoint Similarity) measures the similarity between ground truth poses and predicted poses. We report standard mean average precision and recall scores: $AP$ (the mean of AP scores with OKS threshold = [0.5:0.95:0.05]), $AP^{50}$, $AP^{75}$, $AP^{M}$ (AP score for instances with medium size), $AP^{L}$ (AP score for instances with large size), $AR$ for a fair comparison with start-of-the-arts.

\subsection{Comparison with the State-of-the-Arts}

\begin{figure}
    \setlength{\abovecaptionskip}{0cm}
    \setlength{\belowcaptionskip}{-0.3cm}
    \centering
    \includegraphics[width=0.8\linewidth]{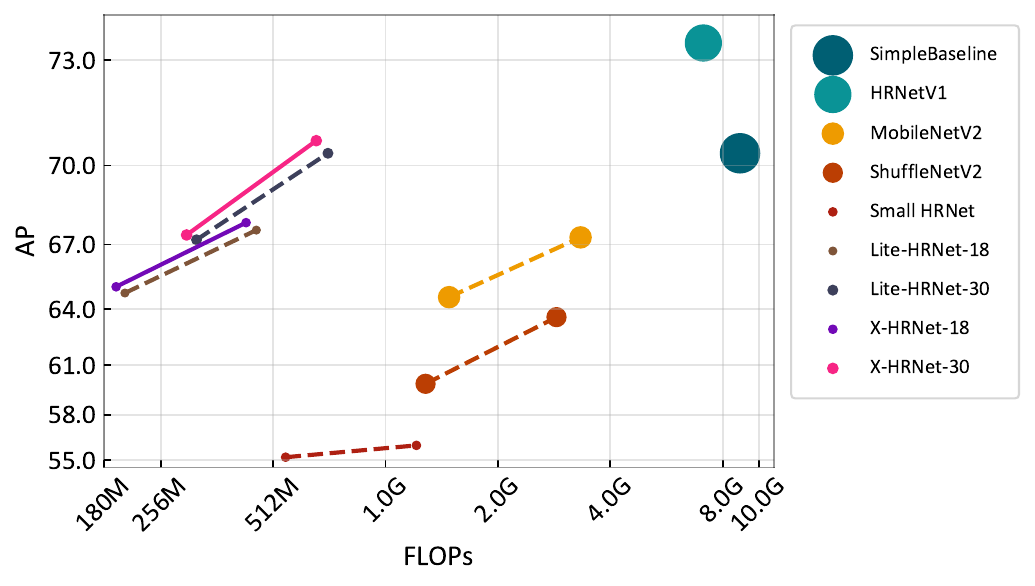}
    \caption{\textbf{Comparisons on the COCO val2017 set.} The size of points denotes the \#Params for the corresponding method. Points with the same color represent the same method trained with different input sizes: $256 \times 192$ and $384 \times 288$. The single points represent those large networks trained with $384 \times 288$. To highlight our method, we draw solid lines for our method and dashed lines for other methods.
    Best viewed in color.}
    \label{fig:experiment-val-performance}
\end{figure}

\noindent\textbf{COCO val2017.}
As illustrated in Figure \ref{fig:experiment-val-performance}, our X-HRNet requires much less FLOPs and achieves competitive performance when compared with other start-of-the-art methods.
When the input size is $256 \times 192$, our X-HRNet-30 achieves a $67.5$ AP score, outperforming existing lightweight methods.
Compared to the dominant lightweight networks designed for classification task: MobileNetV2 \cite{howard2018inverted} and ShuffleNetV2 \cite{ma2018shufflenet}, our X-HRNet achieves superior performance far less computational cost and parameters. When compared with ShuffleNetV2 \cite{ma2018shufflenet} that is initialized by the weights pretrained on the ImageNet dataset \cite{krizhevsky2012imagenet}, the X-HRNet-30 significantly achieves $7.6$ AP gain and only requires $15\%$ parameters and $28\%$ FLOPs.

\begin{table*}[]
    \setlength{\abovecaptionskip}{0cm}
    \setlength{\belowcaptionskip}{-0.5cm}
    \centering
    \caption{\textbf{Comparisons on the COCO val2017 dataset.} \#Params and FLOPs are only calculated for the estimation network, not including those for human detection and keypoint grouping. `pretrain' refers to pretraining the backbone for ImageNet classification \cite{krizhevsky2012imagenet}.}
    \resizebox{0.95\textwidth}{!}{
    \begin{tabular}{@{}llcccccccccc@{}}
        \toprule
        \multicolumn{1}{l|}{model}               & \multicolumn{1}{l|}{backbone}            & \multicolumn{1}{c|}{pretrain} & \multicolumn{1}{c|}{input size} & \multicolumn{1}{c|}{\#Params} & \multicolumn{1}{c|}{FLOPs}  & \multicolumn{1}{c}{$AP$}    & \multicolumn{1}{c}{$AP^{50}$} & \multicolumn{1}{c}{$AP^{75}$} & \multicolumn{1}{c}{$AP^{M}$} & \multicolumn{1}{c}{$AP^{L}$} & $AR$   \\ \specialrule{.05em}{.25ex}{.25ex}
        \multicolumn{12}{l}{Large Networks}                                                                                                                                                                                                                                                                                                                                                                                                          \\ \specialrule{.05em}{.25ex}{.25ex}
        \multicolumn{1}{l|}{SimpleBaseline \cite{xiao2018simple}}      & \multicolumn{1}{l|}{ResNet-50}           & \multicolumn{1}{c|}{Y}        & \multicolumn{1}{c|}{$256 \times 192$}    & \multicolumn{1}{c|}{34.0M}    & \multicolumn{1}{c|}{8.90G}  & \multicolumn{1}{c}{70.4} & \multicolumn{1}{c}{88.6}                   & \multicolumn{1}{c}{78.3}                   & \multicolumn{1}{c}{67.1}                  & \multicolumn{1}{c}{77.2}                  & 76.3 \\
        \multicolumn{1}{l|}{HRNetV1 \cite{wang2020deep}}             & \multicolumn{1}{l|}{HRNetV1-W32}         & \multicolumn{1}{c|}{N}        & \multicolumn{1}{c|}{$256 \times 192$}    & \multicolumn{1}{c|}{28.5M}    & \multicolumn{1}{c|}{7.10G}  & \multicolumn{1}{c}{73.4} & \multicolumn{1}{c}{89.5}                   & \multicolumn{1}{c}{80.7}                   & \multicolumn{1}{c}{70.2}                  & \multicolumn{1}{c}{80.1}                  & 78.9 \\ \specialrule{.05em}{.25ex}{.25ex}
        \multicolumn{12}{l}{Lightweight Classification Networks}                                                                                                                                                                                                                                                                                                                                                                                                    \\ \specialrule{.05em}{.25ex}{.25ex}
        \multicolumn{1}{l|}{MobileNetV2 \cite{howard2018inverted}}         & \multicolumn{1}{l|}{MobileNetV2}         & \multicolumn{1}{c|}{Y}        & \multicolumn{1}{c|}{$256 \times 192$}    & \multicolumn{1}{c|}{9.6M}     & \multicolumn{1}{c|}{1.48G}  & \multicolumn{1}{c}{64.6} & \multicolumn{1}{c}{87.4}                   & \multicolumn{1}{c}{72.3}                   & \multicolumn{1}{c}{61.1}                  & \multicolumn{1}{c}{71.2}                  & 70.7 \\
        \multicolumn{1}{l|}{ShuffleNetV2 \cite{ma2018shufflenet}}        & \multicolumn{1}{l|}{ShuffleNetV2}        & \multicolumn{1}{c|}{Y}        & \multicolumn{1}{c|}{$256 \times 192$}    & \multicolumn{1}{c|}{7.6M}     & \multicolumn{1}{c|}{1.28G}  & \multicolumn{1}{c}{59.9} & \multicolumn{1}{c}{85.4}                   & \multicolumn{1}{c}{66.3}                   & \multicolumn{1}{c}{56.6}                  & \multicolumn{1}{c}{66.2}                  & 66.4 \\
        \multicolumn{1}{l|}{DY-ReLU \cite{chen2020dynamic}}             & \multicolumn{1}{l|}{MobileNetV2}         & \multicolumn{1}{c|}{Y}        & \multicolumn{1}{c|}{$256 \times 192$}    & \multicolumn{1}{c|}{9.0M}     & \multicolumn{1}{c|}{1.03G}  & \multicolumn{1}{c}{68.1} & \multicolumn{1}{c}{88.5}                   & \multicolumn{1}{c}{76.2}                   & \multicolumn{1}{c}{64.8}                  & \multicolumn{1}{c}{74.3}                  & -    \\ \specialrule{.05em}{.25ex}{.25ex}
        \multicolumn{12}{l}{Lightweight Pose Estimation Networks}                                                                                                                                                                                                                                                                                                                                                                                    \\ \specialrule{.05em}{.25ex}{.25ex}
        \multicolumn{1}{l|}{Small HRNet}         & \multicolumn{1}{l|}{HRNet-W16}           & \multicolumn{1}{c|}{N}        & \multicolumn{1}{c|}{$256 \times 192$}    & \multicolumn{1}{c|}{1.3M}     & \multicolumn{1}{c|}{0.54G}  & \multicolumn{1}{c}{55.2} & \multicolumn{1}{c}{83.7}                   & \multicolumn{1}{c}{62.4}                   & \multicolumn{1}{c}{52.3}                  & \multicolumn{1}{c}{61.0}                  & 62.1 \\
        \multicolumn{1}{l|}{Small HRNet}         & \multicolumn{1}{l|}{HRNet-W16}           & \multicolumn{1}{c|}{N}        & \multicolumn{1}{c|}{$384 \times 288$}    & \multicolumn{1}{c|}{1.3M}     & \multicolumn{1}{c|}{1.21G}  & \multicolumn{1}{c}{56.0} & \multicolumn{1}{c}{83.8}                   & \multicolumn{1}{c}{63.0}                   & \multicolumn{1}{c}{52.4}                  & \multicolumn{1}{c}{62.6}                  & 62.6 \\ \specialrule{.05em}{.1ex}{.15ex}

        \multicolumn{1}{l|}{Lite-HRNet \cite{yu2021lite}}       & \multicolumn{1}{l|}{Lite-HRNet-18}       & \multicolumn{1}{c|}{N}        & \multicolumn{1}{c|}{$256 \times 192$}    & \multicolumn{1}{c|}{\textbf{1.1M}}    & \multicolumn{1}{c|}{205.2M} & \multicolumn{1}{c}{64.8} & \multicolumn{1}{c}{86.7}                   & \multicolumn{1}{c}{\textbf{73.0}}                   & \multicolumn{1}{c}{62.1}                  & \multicolumn{1}{c}{70.5}                  & 71.2 \\
        \multicolumn{1}{l|}{X-HRNet (\textbf{Ours})}          & \multicolumn{1}{l|}{X-HRNet-18}          & \multicolumn{1}{c|}{N}        & \multicolumn{1}{c|}{$256 \times 192$}     & \multicolumn{1}{c|}{1.3M}        & \multicolumn{1}{c|}{\textbf{194.4M}} &       \multicolumn{1}{c}{\textbf{65.1}}         & \multicolumn{1}{c}{\textbf{86.7}}       & \multicolumn{1}{c}{72.7}     & \multicolumn{1}{c}{\textbf{62.3}}                       & \multicolumn{1}{c}{\textbf{70.9}}                 & \textbf{71.2}    \\ \specialrule{.01em}{.1ex}{.1ex}

        \multicolumn{1}{l|}{Lite-HRNet}       & \multicolumn{1}{l|}{Lite-HRNet-18}       & \multicolumn{1}{c|}{N}        & \multicolumn{1}{c|}{$384 \times 288$}    & \multicolumn{1}{c|}{\textbf{1.1M}}    & \multicolumn{1}{c|}{461.6M} & \multicolumn{1}{c}{67.6} & \multicolumn{1}{c}{\textbf{87.8}}                   & \multicolumn{1}{c}{75.0}                   & \multicolumn{1}{c}{64.5}                  & \multicolumn{1}{c}{73.7}                  & \textbf{73.7} \\
        \multicolumn{1}{l|}{X-HRNet (\textbf{Ours})}          & \multicolumn{1}{l|}{X-HRNet-18}          & \multicolumn{1}{c|}{N}        & \multicolumn{1}{c|}{$384 \times 288$}    & \multicolumn{1}{c|}{1.3M}         & \multicolumn{1}{c|}{\textbf{433.2M}}       & \multicolumn{1}{c}{\textbf{67.9}}     & \multicolumn{1}{c}{87.6}                       & \multicolumn{1}{c}{\textbf{75.5}}                       & \multicolumn{1}{c}{\textbf{64.7}}                      & \multicolumn{1}{c}{\textbf{73.9}}                      & 73.6 \\ \specialrule{.01em}{.1ex}{.1ex}

        \multicolumn{1}{l|}{Lite-HRNet}       & \multicolumn{1}{l|}{Lite-HRNet-30}       & \multicolumn{1}{c|}{N}        & \multicolumn{1}{c|}{$256 \times 192$}    & \multicolumn{1}{c|}{\textbf{1.8M}}    & \multicolumn{1}{c|}{319.2M} & \multicolumn{1}{c}{67.2} & \multicolumn{1}{c}{\textbf{88.0}}                   & \multicolumn{1}{c}{75.0}                   & \multicolumn{1}{c}{64.3}                  & \multicolumn{1}{c}{73.1}                  & 73.3 \\
        \multicolumn{1}{l|}{X-HRNet (\textbf{Ours})}          & \multicolumn{1}{l|}{X-HRNet-30}          & \multicolumn{1}{c|}{N}        & \multicolumn{1}{c|}{$256 \times 192$}    & \multicolumn{1}{c|}{2.1M}         & \multicolumn{1}{c|}{\textbf{300.2M}}       & \multicolumn{1}{c}{\textbf{67.4}}     & \multicolumn{1}{c}{87.5}                       & \multicolumn{1}{c}{\textbf{75.4}}                       & \multicolumn{1}{c}{\textbf{64.5}}                      & \multicolumn{1}{c}{\textbf{73.3}}                      & \textbf{73.5} \\ \specialrule{.01em}{.1ex}{.1ex}

        \multicolumn{1}{l|}{Lite-HRNet}       & \multicolumn{1}{l|}{Lite-HRNet-30}       & \multicolumn{1}{c|}{N}        & \multicolumn{1}{c|}{$384 \times 288$}    & \multicolumn{1}{c|}{\textbf{1.8M}}    & \multicolumn{1}{c|}{717.8M} & \multicolumn{1}{c}{70.4} & \multicolumn{1}{c}{88.7}                   & \multicolumn{1}{c}{77.7}                   & \multicolumn{1}{c}{67.5}                  & \multicolumn{1}{c}{76.3}                  & \textbf{76.2} \\ 

        \multicolumn{1}{l|}{X-HRNet (\textbf{Ours})}          & \multicolumn{1}{l|}{X-HRNet-30}          & \multicolumn{1}{c|}{N}        & \multicolumn{1}{c|}{$384 \times 288$}    & \multicolumn{1}{c|}{2.1M}         & \multicolumn{1}{c|}{\textbf{668.0M}}       & \multicolumn{1}{c}{\textbf{70.6}}     & \multicolumn{1}{c}{\textbf{88.9}}                       & \multicolumn{1}{c}{\textbf{77.7}}                       & \multicolumn{1}{c}{\textbf{67.6}}                      & \multicolumn{1}{c}{\textbf{76.5}}                      & 76.1 \\ \bottomrule
    \end{tabular}}
    \label{tab:experiment-val-performance}
\end{table*}

As for existing pose estimation approaches, our X-HRNet-18 outperforms the small HRNet by $9.9$ AP points only requiring $35\%$ FLOPs. The performance gap is further enlarged to $11.8$ when the input size is expanded to $384 \times 288$, which demonstrates the superiority of our SUSA module in capturing high-resolution details in our approach.
Regardless of different input sizes or model depths, X-HRNet comprehensively surpasses Lite-HRNet in performance with less FLOPs.
Compared with the large pose estimation network, SimpleBaseline \cite{xiao2018simple}, the X-HRNet-30 significantly reduces the complexity by $97\%$ but merely a $4\%$ performance drop.

\vspace{4pt}\noindent\textbf{COCO test-dev2017.}
With the input size $384 \times 288$, our X-HRNet-18, X-HRNet-30 respectively achieve 67.3 and 70.0 AP scores on COCO test-dev2017 dataset, as reported in Table \ref{tab:experiment-test-performance}. When compared with the Lite-HRNet, our X-HRNet maintains the superiority, which demonstrates the robustness of our X-HRNet.
Compared to large networks, our X-HRNet-30 also achieves comparable performance but requires far less FLOPs and parameters.

\begin{table}[]
    \setlength{\abovecaptionskip}{0.1cm}
    \setlength{\belowcaptionskip}{-0.3cm}
    \centering
    \caption{\textbf{Comparisons on the COCO test-dev2017 dataset.} The input size is set as $384 \times 288$. \#Params and FLOPs are only calculated for the estimation network, not including those for human detection and keypoint grouping.}
    \resizebox{\linewidth}{!}{%
    \begin{tabular}{@{}llcccccc@{}}
    \toprule
    \multicolumn{1}{l|}{model}                                 & \multicolumn{1}{l|}{backbone}      & \multicolumn{1}{c|}{\#Params} & \multicolumn{1}{c|}{FLOPs}  & \multicolumn{1}{c}{$AP$} & \multicolumn{1}{c}{$AP^{50}$} & \multicolumn{1}{c}{$AP^{75}$} & $AR$ \\ \specialrule{.05em}{.25ex}{.25ex}
    \multicolumn{8}{l}{Large Networks}                                                                                                                                                                                                                             \\ \specialrule{.05em}{.25ex}{.25ex}
    \multicolumn{1}{l|}{SimpleBaseline \cite{xiao2018simple}}  & \multicolumn{1}{l|}{ResNet-152}    & \multicolumn{1}{c|}{68.6M}    & \multicolumn{1}{c|}{35.6G}  & \multicolumn{1}{c}{73.7} & \multicolumn{1}{c}{91.9}      & \multicolumn{1}{c}{81.1}     & 79.0 \\
    \multicolumn{1}{l|}{HRNetV1 \cite{wang2020deep}}           & \multicolumn{1}{l|}{HRNetV1-w48}   & \multicolumn{1}{c|}{63.6M}    & \multicolumn{1}{c|}{32.9G}  & \multicolumn{1}{c}{75.5} & \multicolumn{1}{c}{92.5}      & \multicolumn{1}{c}{83.3}     & 80.5 \\ \specialrule{.05em}{.25ex}{.25ex}
    \multicolumn{8}{l}{Lightweight networks}                                                                                                                                                                                                                       \\ \specialrule{.05em}{.25ex}{.25ex}
    \multicolumn{1}{l|}{MobileNetV2 \cite{howard2018inverted}} & \multicolumn{1}{l|}{MobileNetV2}   & \multicolumn{1}{c|}{9.6M}     & \multicolumn{1}{c|}{3.33G}  & \multicolumn{1}{c}{66.8} & \multicolumn{1}{c}{90.0}      & \multicolumn{1}{c}{74.0}     & 72.3 \\
    \multicolumn{1}{l|}{ShuffleNetV2 \cite{ma2018shufflenet}}  & \multicolumn{1}{l|}{ShuffleNetV2}  & \multicolumn{1}{c|}{7.6M}     & \multicolumn{1}{c|}{2.87G}  & \multicolumn{1}{c}{62.9} & \multicolumn{1}{c}{88.5}      & \multicolumn{1}{c}{69.4}     & 68.9 \\
    \multicolumn{1}{l|}{Small HRNet}                           & \multicolumn{1}{l|}{HRNet-16}      & \multicolumn{1}{c|}{1.3M}     & \multicolumn{1}{c|}{1.21G}  & \multicolumn{1}{c}{55.2} & \multicolumn{1}{c}{85.8}      & \multicolumn{1}{c}{61.4}     & 61.5 \\ \specialrule{.01em}{.1ex}{.1ex}
    \multicolumn{1}{l|}{Lite-HRNet \cite{yu2021lite}}          & \multicolumn{1}{l|}{Lite-HRNet-18} & \multicolumn{1}{c|}{\textbf{1.1M}}     & \multicolumn{1}{c|}{461.6M} & \multicolumn{1}{c}{66.9} & \multicolumn{1}{c}{89.4}      & \multicolumn{1}{c}{74.4}     & 72.6 \\
    \multicolumn{1}{l|}{X-HRNet (\textbf{Ours})}                & \multicolumn{1}{l|}{X-HRNet-18}    & \multicolumn{1}{c|}{1.3M}     & \multicolumn{1}{c|}{\textbf{433.2M}} & \multicolumn{1}{c}{\textbf{67.3}} & \multicolumn{1}{c}{\textbf{89.8}}      & \multicolumn{1}{c}{\textbf{74.8}}     & \textbf{73.0} \\ \specialrule{.01em}{.1ex}{.1ex}
    \multicolumn{1}{l|}{Lite-HRNet}                            & \multicolumn{1}{l|}{Lite-HRNet-30} & \multicolumn{1}{c|}{\textbf{1.8M}}     & \multicolumn{1}{c|}{717.8M} & \multicolumn{1}{c}{69.7} & \multicolumn{1}{c}{\textbf{90.7}}      & \multicolumn{1}{c}{77.5}     & 75.4 \\
    \multicolumn{1}{l|}{X-HRNet (\textbf{Ours})}                & \multicolumn{1}{l|}{X-HRNet-30}    & \multicolumn{1}{c|}{2.1M}     & \multicolumn{1}{c|}{\textbf{668.0M}} & \multicolumn{1}{c}{\textbf{70.0}} & \multicolumn{1}{c}{90.6}     & \multicolumn{1}{c}{\textbf{77.7}}     & \textbf{75.5} \\ \bottomrule
    \end{tabular}
    }
    \label{tab:experiment-test-performance}
    \end{table}

\subsection{Ablation Study}
We perform ablation studies and report the results on the COCO val2017 set. And all results are obtained with the input size of $256 \times 192$.

\vspace{4pt}\noindent\textbf{The effect of SUSA modules.}
To compare our SUSA modules with the costly $1\times1$ convolution, we make the comparison between our X-HRNet and wider naive Lite-HRNet (WNL-HRNet) \cite{yu2021lite} whose main block is the standard shuffle block. And we simply remove two $1\times1$ convolutions in all shuffle blocks and set the resulting network as the baseline. As reported in Table \ref{tab:experiments-ablation-modules}, the WNL-HRNet achieves a $65.7$ AP score, which is decreased by $4.4$ points when removing two $1\times1$ convolutions in shuffle blocks. Meanwhile, the removing also reduces the FLOPs by $122.2$M, and $61.1$ MFLOPs for each $1\times1$ convolution in shuffle block. 
By inserting one H-wise or W-wise SUSA after the depthwise $3\times3$ convolution in each incomplete shuffle block, the baseline is respectively improved by $2.0$, $1.6$. And with both of them (X-HRNet), the performance is consistently increased by $3.8$, which is larger than the sum of AP improvements ($3.6$) of adding either one.
% Compared to the WNL-HRNet, our X-HRNet reduces the complexity by more than $60\%$ and achieves a comparable AP score, $65.1$.
It is worth noting that our SUSA modules only cost $2.7$ MFLOPs that is only $4\%$ of the $1\times1$ convolution but achieves a comparable AP score, $65.1$.
The $1\times1$ convolution plays a significant role in information exchange across channels, but it is computationally costly. Our SUSA module plays a similar role as $1\times1$ convolution by learning cross-shape attention vectors and only needs to perform convolution calculation on a single spatial dimension.

\begin{table}[]
    \setlength{\abovecaptionskip}{0.1cm}
    \setlength{\belowcaptionskip}{-0.3cm}
    \centering
    \caption{\textbf{Ablation study about SUSA modules.} All results are evaluated on COCO val2017 set and the input size is $256 \times 192$. WNL-HRNet-18 = wider naive Lite-HRNet \cite{yu2021lite}, baseline refers to dropping two $1\times1$ convolutions from shuffle blocks in WNL-HRNet-18, + SUSA = inserting the SUSA module into the incomplete shuffle block.}
    \resizebox{\linewidth}{!}{%
    \begin{tabular}{@{}l|c|c|cccc@{}}
    \toprule
    \multicolumn{1}{c|}{model}                      & \#Params & FLOPs   & $AP$   & $AP^{50}$ & $AP^{75}$ & $AR$   \\ \specialrule{.05em}{.25ex}{.25ex}
    \multicolumn{1}{l|}{WNL-HRNet-18  }             & 1.3M     & 311.1M  & $65.7$ & $87.0$    & $73.3$    & $71.8$ \\ \specialrule{.05em}{.25ex}{.25ex}
    \multicolumn{1}{l|}{Baseline          }         & 0.9M     & 188.9M  & $61.3$ & $85.3$    & $68.7$    & $67.7$ \\
    \multicolumn{1}{l|}{+ W-wise SUSA     }         & 1.1M     & 191.2M  & $62.9$ & $85.7$    & $70.5$    & $69.1$ \\
    \multicolumn{1}{l|}{+ H-wise SUSA     }         & 1.1M     & 191.9M  & $63.3$ & $85.9$    & $70.7$    & $69.6$ \\
    \multicolumn{1}{l|}{X-HRNet-18       }          & 1.3M     & 194.4M  & \textbf{65.1} &  \textbf{86.7}     &   \textbf{72.7}   & \textbf{71.2} \\\bottomrule
    \end{tabular}%
    }
    
    \label{tab:experiments-ablation-modules}
    \end{table}

\begin{table}[]
    \setlength{\abovecaptionskip}{0.1cm}
    \setlength{\belowcaptionskip}{-0.3cm}
    \centering
    \caption{\textbf{Ablation study about different fusion types.} All results are evaluated on the COCO val2017 set, and the input size is $256 \times 192$. The `add' denotes addition fusion, and the `mul' refers to multiplication fusion.}
    \resizebox{\linewidth}{!}{%
    \begin{tabular}{@{}l|c|c|cccc@{}}
    \toprule
    \multicolumn{1}{c|}{model}                      & \#Params & FLOPs   & $AP$   & $AP^{50}$ & $AP^{75}$ & $AR$   \\ \specialrule{.05em}{.25ex}{.25ex}
    \multicolumn{1}{l|}{X-HRNet (add)  }          & 1.3M     & 194.4M & $63.3$ & $85.9$    & $70.7$    & $69.6$ \\
    \multicolumn{1}{l|}{X-HRNet (mul)  }          & 1.3M     & 194.4M  & \textbf{65.1} &  \textbf{86.7}     &   \textbf{72.7}   & \textbf{71.2} \\\bottomrule
    \end{tabular}%
    }
    \label{tab:experiments-ablation-fusion}
    \end{table}

\vspace{4pt}\noindent\textbf{The effect of different fusion types.}
The different choices on fusion are influential for the SUSA module to achieve high performance. And we respectively apply addition and multiplication fusion on the GC block and our SUSA module to verify our analysis in \ref{relation-to-gc-block}. Ablation results on different fusion are shown in Table \ref{tab:experiments-ablation-fusion}. The experimental results are consistent with our aforementioned theoretical analysis. The multiplication fusion is more suitable for the human pose estimation task than addition fusion, which is inverse for object detection or segmentation as illustrated in \cite{cao2019gcnet}.

\begin{table}[]
    \setlength{\abovecaptionskip}{0.1cm}
    \setlength{\belowcaptionskip}{-0.3cm}
    \centering
    \caption{\textbf{Ablation study about the arrangement order of the H-wise and W-wise SUSA.} $^{\intercal}$ indicates the arrangement order of the H-wise and W-wise SUSA in the X-shuffle block is reversed.}
    \resizebox{\linewidth}{!}{
    \begin{tabular}{@{}c|c|l|llll@{}}
    \toprule
    model                 & \#Params & FLOPs & $AP$ & $AP^{50}$ & $AP^{75}$ & $AR$ \\ \specialrule{.05em}{.25ex}{.25ex}
    \multicolumn{1}{l|}{X-HRNet-18$^{\intercal}$} & 1.3M    & 194.4M & 64.5 &  86.4     &   72.3    & 70.8 \\ 
    \multicolumn{1}{l|}{X-HRNet-18}           & 1.3M    & 194.4M & \textbf{65.1} &  \textbf{86.7}     &   \textbf{72.7}   & \textbf{71.2} \\\bottomrule
    \end{tabular}}
    
    \label{tab:experiments-ablation-HWorWH}
    \end{table} 

\vspace{4pt}\noindent\textbf{The effect of the order of H-wise and W-wise SUSA.}
As mentioned in Sec. \ref{method-X-HRNet}, the order of H-wise and W-wise SUSA in the X-shuffle block has a surprising influence on performance. The X-HRNet outperforms the reversed one by $0.7$ AP points, and they have the same \#Params and FLOPs, as shown in Table \ref{tab:experiments-ablation-HWorWH}.
This peculiar discrepancy is caused by the depthwise $3\times3$ convolution between two SUSA modules, which enlarges the receptive field for the second SUSA. And for the COCO benchmark dataset, the prescriptive input size is $256 \times 192$ or $384 \times 288$, where height is larger than width. Consequently, the X-shuffle block whose H-wise SUSA is equipped with the larger receptive field outperforms the reversed one. 

\section{Conclusion}
This paper presents a lightweight human pose estimation approach named X-HRNet. 
We explore the essential characteristic of the pose estimation task: estimating human joints by 2D single-peak heatmaps, where each 2D heatmap can be horizontally and vertically projected to and reconstructed by a pair of 1D heat vectors.
Inspired by this characteristic, we propose a lightweight module named Spatially Unidimensional Self-Attention (SUSA). The resulting lightweight pose estimation network X-HRNet achieves the state-of-the-art in terms of complexity and accuracy trade-off on the COCO benchmark.

\bibliographystyle{IEEEbib}
\small
\bibliography{icme2022template}

\begin{thebibliography}{10}

\bibitem{badrinarayanan2017segnet}
Vijay Badrinarayanan et~al.,
\newblock ``Segnet: A deep convolutional encoder-decoder architecture for image segmentation,''
\newblock {\em Trans.PAMI}, vol. 39, no. 12, pp. 2481--2495, 2017.

\bibitem{chen2014semantic}
Liang-Chieh Chen et~al.,
\newblock ``Semantic image segmentation with deep convolutional nets and fully connected crfs,''
\newblock {\em ICLR}, 2014.

\bibitem{sun2019deep}
Ke~Sun et~al.,
\newblock ``Deep high-resolution representation learning for human pose estimation,''
\newblock in {\em Proceedings of the IEEE/CVF Conference on Computer Vision and Pattern Recognition}, 2019, pp. 5693--5703.

\bibitem{wang2020deep}
Jingdong Wang et~al.,
\newblock ``Deep high-resolution representation learning for visual recognition,''
\newblock {\em Trans.PAMI}, 2020.

\bibitem{cheng2020higherhrnet}
Bowen Cheng et~al.,
\newblock ``Higherhrnet: Scale-aware representation learning for bottom-up human pose estimation,''
\newblock in {\em CVPR}, 2020, pp. 5386--5395.

\bibitem{li2020simple}
Jia Li et~al.,
\newblock ``Simple pose: Rethinking and improving a bottom-up approach for multi-person pose estimation,''
\newblock in {\em AAAI}, 2020, pp. 11354--11361.

\bibitem{cao2019gcnet}
Yue Cao et~al.,
\newblock ``Gcnet: Non-local networks meet squeeze-excitation networks and beyond,''
\newblock in {\em CVPR}, 2019.

\bibitem{zhang2018shufflenet}
Xiangyu Zhang et~al.,
\newblock ``Shufflenet: An extremely efficient convolutional neural network for mobile devices,''
\newblock in {\em CVPR}, 2018, pp. 6848--6856.

\bibitem{ma2018shufflenet}
Ningning Ma et~al.,
\newblock ``Shufflenet v2: Practical guidelines for efficient cnn architecture design,''
\newblock in {\em ECCV}, 2018, pp. 116--131.

\bibitem{howard2018inverted}
Andrew Howard et~al.,
\newblock ``Inverted residuals and linear bottlenecks: Mobile networks for classification, detection and segmentation,''
\newblock in {\em CVPR}, 2018.

\bibitem{chollet2017xception}
Fran{\c{c}}ois Chollet,
\newblock ``Xception: Deep learning with depthwise separable convolutions,''
\newblock in {\em CVPR}, 2017, pp. 1251--1258.

\bibitem{howard2017mobilenets}
Andrew~G Howard et~al.,
\newblock ``Mobilenets: Efficient convolutional neural networks for mobile vision applications,''
\newblock {\em arXiv preprint arXiv:1704.04861}, 2017.

\bibitem{yu2021lite}
Changqian Yu et~al.,
\newblock ``Lite-hrnet: A lightweight high-resolution network,''
\newblock in {\em CVPR}, 2021, pp. 10440--10450.

\bibitem{yu2018bisenet}
Changqian Yu et~al.,
\newblock ``Bisenet: Bilateral segmentation network for real-time semantic segmentation,''
\newblock in {\em ECCV}, 2018, pp. 325--341.

\bibitem{lin2017feature}
Tsung-Yi Lin et~al.,
\newblock ``Feature pyramid networks for object detection,''
\newblock in {\em CVPR}, 2017, pp. 2117--2125.

\bibitem{xiao2018simple}
Bin Xiao et~al.,
\newblock ``Simple baselines for human pose estimation and tracking,''
\newblock in {\em ECCV}, 2018, pp. 466--481.

\bibitem{chen2018cascaded}
Yilun Chen et~al.,
\newblock ``Cascaded pyramid network for multi-person pose estimation,''
\newblock in {\em CVPR}, 2018, pp. 7103--7112.

\bibitem{wang2018non}
Xiaolong Wang et~al.,
\newblock ``Non-local neural networks,''
\newblock in {\em CVPR}, 2018, pp. 7794--7803.

\bibitem{lin2014microsoft}
Tsung-Yi Lin et~al.,
\newblock ``Microsoft coco: Common objects in context,''
\newblock in {\em ECCV}. Springer, 2014, pp. 740--755.

\bibitem{krizhevsky2012imagenet}
Alex Krizhevsky et~al.,
\newblock ``Imagenet classification with deep convolutional neural networks,''
\newblock {\em NeurIPS}, vol. 25, pp. 1097--1105, 2012.

\bibitem{chen2020dynamic}
Yinpeng Chen et~al.,
\newblock ``Dynamic relu,''
\newblock in {\em ECCV}. Springer, 2020, pp. 351--367.

\end{thebibliography}

\end{document}